\documentclass[letterpaper, 10 pt, conference]{ieeeconf}  

\IEEEoverridecommandlockouts                              

\usepackage{cite}
\usepackage{amsmath,amssymb,amsfonts}

\usepackage{graphicx}
\usepackage{textcomp}
\usepackage{xcolor}

\usepackage{algorithm}
\usepackage{algpseudocode}

\newcommand{\rhat}{\hat{R}}

\title{\LARGE \bf
Advice Conformance Verification by Reinforcement
Learning agents for Human-in-the-Loop
}

\author{Mudit Verma$^{1}$, Ayush Kharkwal$^{1}$, and Subbarao Kambhampati$^{1}$
\thanks{$^{1}$SCAI, Arizona State University, AZ , 85281}
}

\begin{document}

\maketitle
\thispagestyle{empty}
\pagestyle{empty}

\begin{abstract}

Human-in-the-loop (HiL) reinforcement learning is gaining traction in domains with large action and state spaces, and sparse rewards by allowing the agent to take advice from HiL. Beyond advice accommodation, a sequential decision-making agent must be able to express the extent to which it was able to utilize the human advice. Subsequently, the agent should provide a means for the HiL to inspect parts of advice that it had to reject in favor of the overall environment objective. We introduce the problem of Advice-Conformance Verification which requires reinforcement learning (RL) agents to 
provide assurances to the human in the loop regarding how much of their advice is being conformed to. We then propose a Tree-based lingua-franca to support this communication, called a Preference Tree. We study two cases of good and bad advice scenarios in MuJoCo's Humanoid environment. Through our experiments, we show that our method can provide an interpretable means of solving the Advice-Conformance Verification problem by conveying whether or not the agent is using the human's advice. Finally, we present a human-user study with 20 participants that validates our method.

\end{abstract}

\section{Introduction}
\label{sec:intro}

Deep Reinforcement Learning has struggled with sparse reward environments resulting in several frameworks utilizing Human-in-the-Loop (HiL) that have shown promising success. These works \cite{hu2020learning, expand, fresh, dqntamer, tamer, pebble, christiano} utilize advice or preferences from humans as a form of guidance, however, a missing aspect of these works is the inability of the RL agent to provide assurances to the human user regarding to what extent their advice was accommodated by the agent. We term this problem as the \textbf{Advice-Conformance Verification} problem which requires an RL agent to provide assurances or explanations that conveys whether the agent conforms to the human advice and how much of it the agent let go for the larger interest of completing the task. 

It is well known in the field of Human-aware-AI that humans can form expectations of the agents they are interacting with via several means \cite{trust1, trust2}, for example when they observe the agent's behavior. Similarly, we posit that when an agent requests human advice to achieve the task as determined by environment rewards, the human in the loop may establish the belief that the agent's success on the task is the consequence of following their advice. We leverage \cite{hu2020learning} for observing that so long the agent attempts to optimize for the underlying environment reward, even in the presence of bad advice the agent is still able to obtain a good policy (as shown in Fig. \ref{fig:score}). However, such a belief may be ill-placed (in the event of either a poor advice or misspecified environment rewards) and in many situations for example, where the safety of the human in the loop is of concern, such beliefs should be corrected. The Advice-Conformance Verification captures this issue by requiring agents to allow the HiL to inspect whether the advice was utilized in the intended manner and if possible what parts of the given advice were rejected by the agent.

\section{Background}

In the Human in the Loop Reinforcement learning works like \cite{expand, dqntamer,fresh, hu2020learning}, the learning paradigm involves the agent acting in an environment $\mathcal{E}$ by sensing an observation $o_t \in \mathcal{O}$ at time $t$. As in traditional reinforcement learning these methods model the environment as an MDP tuple $(\mathcal{O, T, A, R})$ where $\mathcal{O, A}$ are the agent's observation and action spaces, $\mathcal{T}$ is the transition function governed by the environment dynamics and $R$ are the environment rewards. Additionally, several works take into account human advice in different ways for example action advice \cite{actionadvising}, policy advice \cite{policyadvice}, or reward advice \cite{hu2020learning, fresh}. We are interested in leveraging works that perform reward shaping \cite{rewardshaping} 
as a means to accommodate human advice. 

The agent's aim in this class of problems is to come up with a policy $\pi_{\theta}$ such that it achieves the maximum possible returns computed over rewards $R$. Note that the agent typically would at least have the environment reward $R$ and a shaped reward $\rhat$ that it computes using the human advice (which itself could be represented in the form of a reward function, say $\mathcal{F}$). The human advice, therefore, is meant to aid the agent in achieving the task specified by rewards $R$. In this work, we will take \cite{hu2020learning} as the backbone HiL RL algorithm and propose a solution to the advice conformance verification problem in this setup.

\begin{figure*}[h]
\begin{center}
\centerline{\includegraphics[width=1.8\columnwidth]{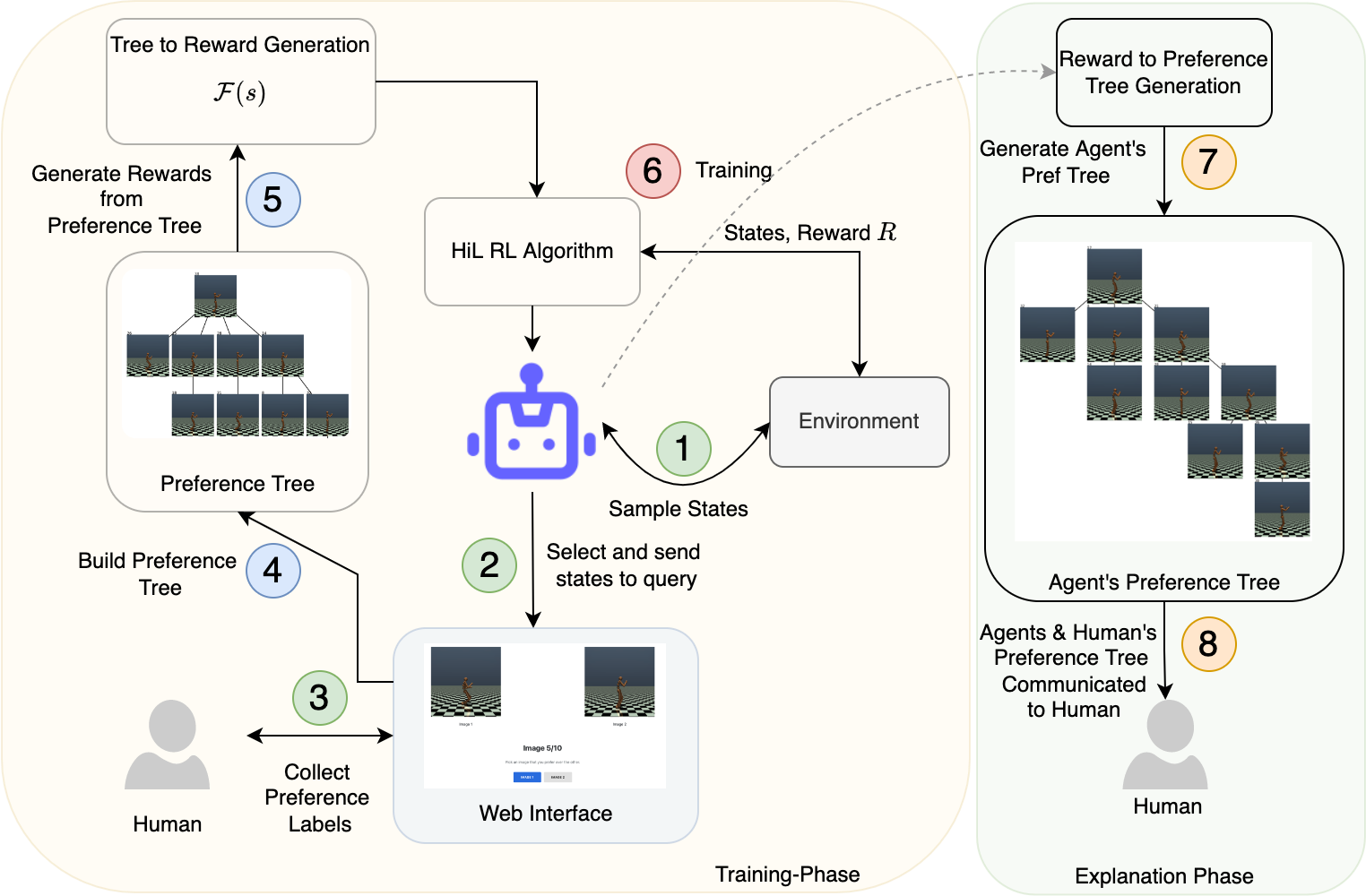}}
\caption{Overview of the Advice Conformance Verification Process}
\label{fig:overview}
\end{center}
\vspace{-1cm}
\end{figure*}

\section{Method}
Our solution to the Advice-Conformance Verification problem is to establish the lingua franca between the human user and the RL agent in the form of a Preference Tree which is a directed acyclic graph computed from the given preferences. The Preference Tree computed using the Human in the Loop is termed the Human-Preference Tree. We present a method to extract a Preference Tree from the RL agent at any time during the agent's training regime, referred to as the Agent's Preference Tree. We use the pair of preference trees, the one extracted by the agent from the human and the other generated by the agent, as a means to convey how the advice has been utilized by the agent in the learning process. We expect a significant deviation of the Agent's Preference tree structure from the Human's Preference Tree to imply a deviation from the human advice. 


\textbf{Sampling Candidate states for the Preference Tree: } For this work, we will use state-based binary preference between two observations to get the tuple $\mathcal{H} = \{(o_i, o_j, b_ij)_k\}$ where $o_i, o_j$ are image observation shown to the user for underlying MDP states $s_i, s_j$ and the binary label $b_i \in \{0,1\}$ where 0 implies human preference of $o_i$ over $o_j$ and vice versa for 1. $k$ such preference labels are collected. The choice of $o_i, o_j, k$ could be domain-dependent, but typically works like \cite{christiano, pebble} have obtained $o_i, o_j$ via uniform sampling techniques (figure \ref{fig:overview} Step 1). As will be discussed in Section \ref{sec:discussion}, better state clustering and exploration techniques are potential methods of extending our work for better interpretable Preference Trees.

\textbf{Creating a Single-Elimination Tournament Tree with HiL:} Since the human is queried pairs of states, we must create the initial pairs from the candidate states. We use the Braverman–Mossel Noisy sampling model \cite{bmmodel2, bmmodel} to create a single-elimination tournament tree, where each candidate state is treated as a ``player" and its corresponding environment reward as the player's intrinsic ability. The state which ends up being preferred by the human can be treating as that state ``beating" the alternative. To allow for the possibility that weaker states can beat stronger players the BM model allows, as input, a probability $p$ where a high value of $p$ entails a random tournament. The ``winner" states are further pitched against each other as a query to the HiL and this process continues until we get a single winner state. Theory on single-elimination tournaments also points to ``rigged" tournaments \cite{bmmodel} where there does exist an initial pairing such that a given state would win the tournament, however, we curtail this issue with a high $p$ value to obtain a random tournament tree. 

\begin{figure}[h]
\begin{center}
\centerline{\includegraphics[width=0.8\columnwidth]{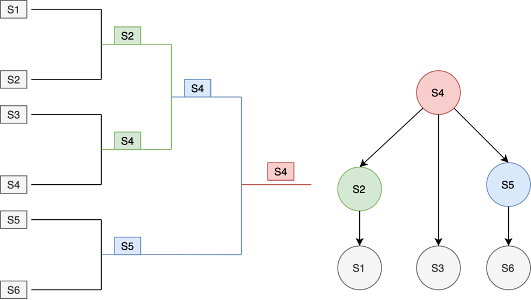}}
\caption{(Best viewed in color) An example of a dendrogram to preference tree construction with the dendrogram tree on the left, where the tournament starts with the given pairings for players $s_1$ through $s_6$. The tree on the right is the constructed Preference tree which condenses the dendrogram by eliminating redundant nodes.}
\label{fig:dendro2pref}
\end{center}
\end{figure}

\textbf{Conversion of Single-Elimination Tournament Tree to a Preference Tree :} From the obtained single-elimination tournament dendrogram we use the algorithm \ref{alg:dend2pref} to convert it into a Preference Tree. A preference tree can be thought of as a condensed tournament tree where no nodes are repeated. Figure \ref{fig:dendro2pref} shows an example of dendrogram to preference tree conversion. Algorithm \ref{alg:dend2pref} achieves this in a standard depth-first search recursive manner. In a preference tree, the edge from a node $i$ to $j$ has a weight of $w_{ij} = height(i)-height(j)$ where $height(x)$ is computed in the dendrogram (leaves have a height of 0).

\begin{algorithm}
\caption{Dendrogram to Preference Tree}
\label{alg:dend2pref}
\begin{algorithmic}

\Procedure{recurse}{$head$, Preference Tree $\mathcal{T} = \{head\}$}
\While{$child$ in $head.child$}
\If {$child.value$ is $head.value$}
    \State \Call{recurse}{$child$} \Comment{Skip}
\Else 
    \State $\mathcal{T}.head.children \gets$ $NewNode(child)$
    \State  \Call{recurse}{$child$} 
            \Comment{Add new node to $\mathcal{T}$}
\EndIf
\EndWhile
\State \Return $\mathcal{T}$ \Comment{Constructed Preference Tree}
\EndProcedure
\end{algorithmic}
\end{algorithm}

\textbf{Utilizing the Preference tree for Advice Accommodation :} 
We extend \cite{hu2020learning} to utilize this preference tree as the given preference reward function $\mathcal{F}$ by grounding the preference tree nodes to reward values using the equation \ref{eq:tree2reward}, where $\mathcal{T}$ is the constructed Preference-Tree, $r_e$ is a positive constant (edge reward), $r_b$ is a positive constant (reward to leaves), $w_{sc}$ is the edge weight from node $s$ to child $c$ in $\mathcal{T}$, $\mathcal{T}_r$ is the reward computed for a node $c$ in a previous step of this bottom up method and $\mathcal{C}(s)$ is the set of children of node $s$. 
\begin{equation}
\label{eq:tree2reward}
    \mathcal{T}_r(s) =\begin{cases}
		-r_b, &   \text{if } s \in \mathcal{T}.leaves \\
            \dfrac{1}{|\mathcal{C}(s)|} \displaystyle \sum_{c \sim \mathcal{C}(s)}\mathcal{T}_r(c) + r_e * w_{sc}, & \text{otherwise}
		 \end{cases}
\end{equation}

To assign preference rewards to states absent in the Preference tree, we use a similarity measure like cosine similarity ($d_{sim}$) between state observations. Hence, preference reward for a state, for some $ s' \in \mathcal{T}.nodes$, is computed as : 
\begin{equation}
    \mathcal{F}(s) =  
            \displaystyle \min_{s'} d_{sim}(s,s')*\mathcal{T}_{r}(s')*\mathcal{T}_d(s')
\end{equation}

Where $\mathcal{T}_d(s')$ is the depth of state $s'$ in the Human Preference-Tree.

\textbf{Obtaining Agent Preference Tree :}
Since our approach is ``anytime" we use the agent's shaped reward function $\rhat$ to provide a preference on a pair of states ${s_i, s_j}$ as, 
\begin{equation}
    b_{ij} = \begin{cases}
        & 1,  \mathcal{T}_r(s_i) > \mathcal{T}_r(s_j) \\
        & 0, \text{otherwise}
    \end{cases}
\end{equation}
Therefore, we start with the same initial pairings in our single-elimination tournament constructed using the BM model, however, use our agent's shaped reward function $\Tilde{R}(s, a) = R(s,a) + z_{\phi}(s)\mathcal{F}(s)$  to specify the winner states and construct the Agent's Preference Tree.

\section{Experiments}

For our experiments, we focus on the MuJoCo environment Humanoid-V2 \cite{mujoco} where the optimal policy on environment rewards allow the humanoid to walk. We take two cases of human advice, 
\begin{enumerate}
    \item Case 1 / Good advice : Preference is exactly aligned with environment rewards $R$, i.e. $\mathcal{F}(s) = R(s) \forall s \in \mathcal{S}$. 
    
    \item Case 2 / Bad advice: Human prefers the robot in ``standing position". Note that this piece of advice maybe valid in human's mental model, however does not align with the environment rewards which is why it is termed as Bad advice. We collect data from human in the loop for their preferences using a web interface as in figure \ref{fig:overview}. 
\end{enumerate}

\begin{figure}[h]
\begin{center}
\centerline{\includegraphics[width=\columnwidth]{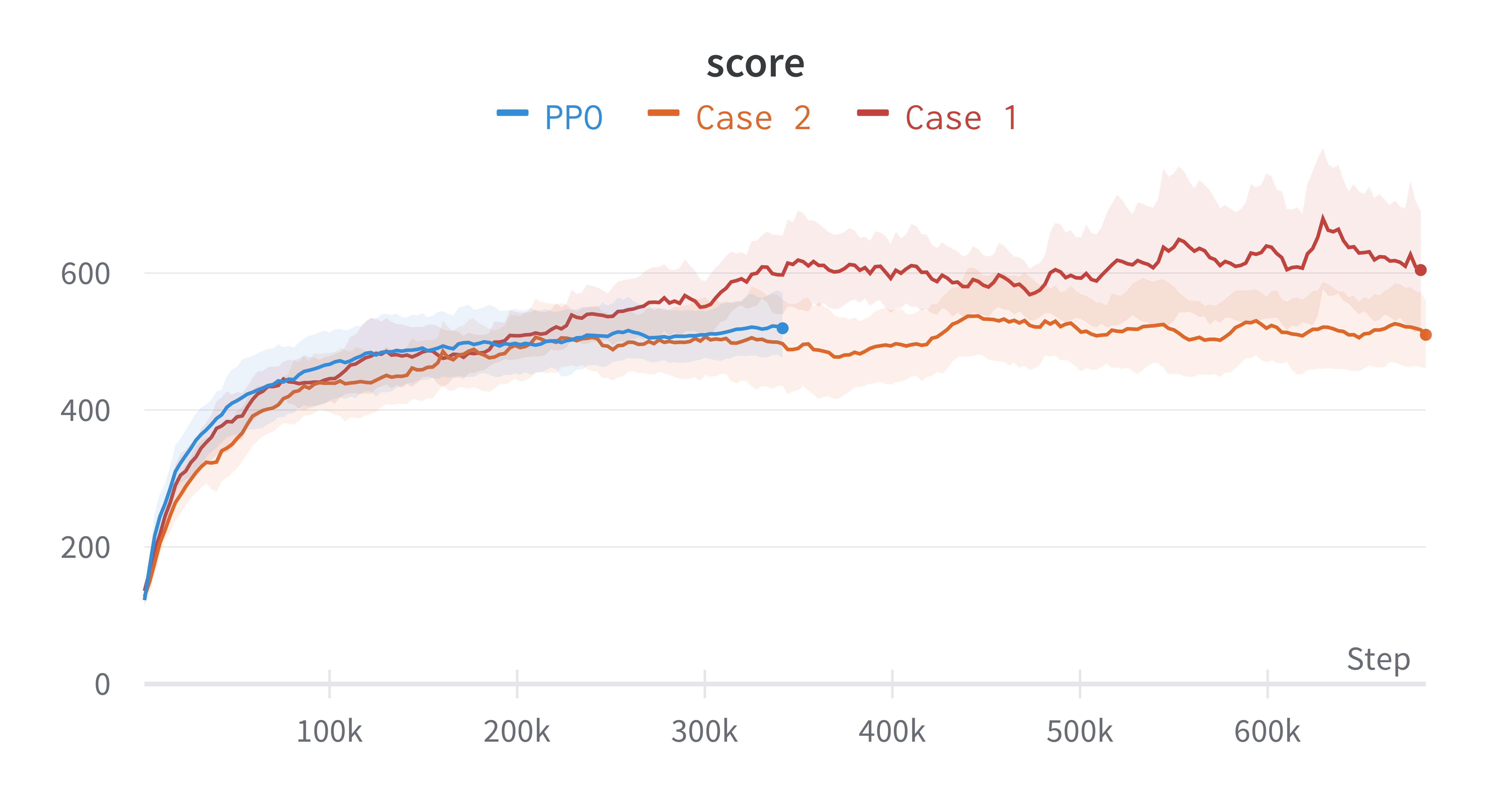}}
\caption{Plot showing the final cumulative reward achieved by 3 agents, PPO agent trained on environment rewards, Case 1 modified agent \cite{hu2020learning} on good advice, and Case 2 modified agent \cite{hu2020learning} on bad advice for MuJoCo's Humanoid-v2. Note that the PPO stops at the 350k mark because the threshold for convergence was met.}
\label{fig:score}
\end{center}
\end{figure}

In Case 1, since the given advice is aligned with the environment rewards, in the shaped reward $\rhat(s) = R(s) + z_{\phi}\mathcal{F}(s)$ and $\mathcal{F}(s) = R(s)$, we find that the learnt $z_\phi$ (using \cite{hu2020learning}) tends to be a constant value across all states thereby retaining all the preference orderings $b_{ij}$, which is what one expects with good-advice. As a consequence, the preference tree generated by the agent in Case 1 correctly matches the Human Preference tree. This shows the effectiveness of Preference Trees in correctly identifying whether the agent is using the advice in the intended manner for aligned advice.

\begin{figure*}[ht]
\begin{center}
\centerline{\includegraphics[width=1.8\columnwidth]{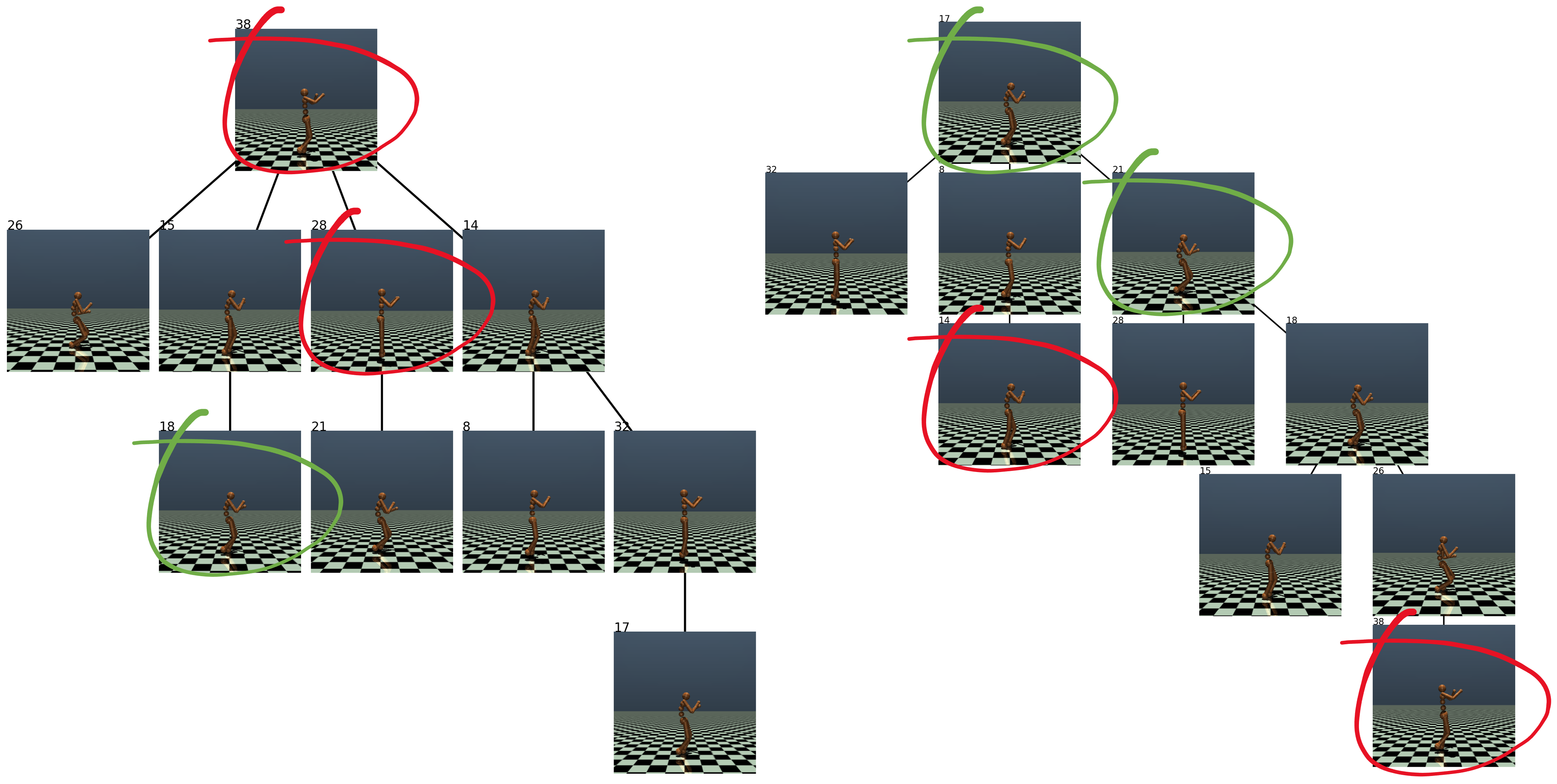}}
\caption{ (Best viewed in color) Preference Trees generated for Case 2. The red and green highlights correspond to states where the humanoid is in bent and straight posture respectively (as determined by the human in the loop). The tree on the left is the Human Preference Tree indicating a human's preference for straight posture positions of the humanoid. On the right, we extract the Agent's Preference Tree at convergence. We find that the agent's shaped rewards value bent postures more than straight postures. These trees (without the highlights) were shown to the participants in our human study.}
\label{fig:combinedresult}
\end{center}
\end{figure*}


In Case 2, since the performance of the agent reaches close to optimal as shown in figure \ref{fig:score}, given just this performance measure, the human might believe that their advice was incorporated and the agent prefers ``standing posture". As discussed before, this can cause the human to develop false expectations on the agent's behavior as well assign false credit to the effectiveness of their advice. When we apply our method to generate the two Preference trees - the Human Preference Tree and the Agent's Preference tree, we find that the agent in fact prefers the ``bent postures" over the `straight posture of ``standing position" as shown in figure \ref{fig:combinedresult}. Hence, we can clearly recognize that the agent had to reject the human advice, a fact that was not reflected without the use of Preference Trees.

We are also interested in recognizing the effectiveness of our explanation strategy via Preference Trees. We conducted a human study with 20 participants to check two hypotheses. First, Can the human correctly recognize that the two trees shown to them imply different preferences? This study attempts to verify our claim that the use of Preference trees is a good means to realize that there is a deviation from the given Human preferences. All of the users (100\%) were able to identify and conclude that the trees convey different preferences. Second, we wanted to verify our claim that Preference Trees can help the human user characterize the deviation. We queried the participants with the two Preference trees and an accompanying text that the first Tree prefers ``agent should prefer a straight posture". We asked them whether they believed that the second tree prefers a ``bent posture". 85\% of the users found that the second tree prefers ``bent postures" hence bolstering our claim that Preference Trees aids in characterizing the agent's deviation from human advice.

\vspace{-0.2cm}
\section{Discussion}
\label{sec:discussion}
We presented the problem of Advice Conformance Verification a solution to which requires a Reinforcement Learning agent to provide a means for the Human in the Loop to inspect whether the agent is utilizing the given advice in the intended manner. We did so using the proposed Preference Trees and showed how they can be used with the HiL work proposed in \cite{hu2020learning}. We ran an experiment on MuJoCo's Humanoid-v2 environment with good and bad advice. We conducted a user study to verify that Preference Trees are a good means to identify whether the agent's shaped reward function deviates from the Human specified preferences and further characterize the possible deviation. 

Our initial results on obtaining Preference trees on better, more diverse states have been very promising. Our future work includes the use of clustering techniques to query diverse states to the human in the loop such that the obtained Preference trees are even more interpretable. We also plan on conducting extensive evaluations on other domains, differing advice, and HiL algorithms that operationalize human advice accommodation via reward shaping techniques.

\bibliographystyle{plain}
\bibliography{bib.bib}

\begin{thebibliography}{10}

\bibitem{dqntamer}
Riku Arakawa, Sosuke Kobayashi, Yuya Unno, Yuta Tsuboi, and Shin-ichi Maeda.
\newblock Dqn-tamer: Human-in-the-loop reinforcement learning with intractable
  feedback.
\newblock {\em arXiv preprint arXiv:1810.11748}, 2018.

\bibitem{bmmodel2}
Mark Braverman and Elchanan Mossel.
\newblock Sorting from noisy information.
\newblock {\em arXiv preprint arXiv:0910.1191}, 2009.

\bibitem{christiano}
Paul~F Christiano, Jan Leike, Tom Brown, Miljan Martic, Shane Legg, and Dario
  Amodei.
\newblock Deep reinforcement learning from human preferences.
\newblock {\em Advances in neural information processing systems}, 30, 2017.

\bibitem{expand}
Lin Guan, Mudit Verma, Suna~Sihang Guo, Ruohan Zhang, and Subbarao Kambhampati.
\newblock Widening the pipeline in human-guided reinforcement learning with
  explanation and context-aware data augmentation.
\newblock {\em Advances in Neural Information Processing Systems},
  34:21885--21897, 2021.

\bibitem{hu2020learning}
Yujing Hu, Weixun Wang, Hangtian Jia, Yixiang Wang, Yingfeng Chen, Jianye Hao,
  Feng Wu, and Changjie Fan.
\newblock Learning to utilize shaping rewards: A new approach of reward
  shaping.
\newblock {\em Advances in Neural Information Processing Systems},
  33:15931--15941, 2020.

\bibitem{actionadvising}
Ercument Ilhan, Jeremy Gow, and Diego Perez-Liebana.
\newblock Action advising with advice imitation in deep reinforcement learning.
\newblock {\em arXiv preprint arXiv:2104.08441}, 2021.

\bibitem{tamer}
W~Bradley Knox and Peter Stone.
\newblock Interactively shaping agents via human reinforcement: The tamer
  framework.
\newblock In {\em Proceedings of the fifth international conference on
  Knowledge capture}, pages 9--16, 2009.

\bibitem{pebble}
Kimin Lee, Laura Smith, and Pieter Abbeel.
\newblock Pebble: Feedback-efficient interactive reinforcement learning via
  relabeling experience and unsupervised pre-training.
\newblock {\em arXiv preprint arXiv:2106.05091}, 2021.

\bibitem{rewardshaping}
Andrew~Y Ng, Daishi Harada, and Stuart Russell.
\newblock Policy invariance under reward transformations: Theory and
  application to reward shaping.
\newblock In {\em Icml}, volume~99, pages 278--287, 1999.

\bibitem{bmmodel}
Isabelle Stanton and Virginia~Vassilevska Williams.
\newblock Manipulating single-elimination tournaments in the braverman-mossel
  model.
\newblock In {\em Workshop on Social Choice and Artificial Intelligence},
  volume~87, 2011.

\bibitem{policyadvice}
Kaushik Subramanian, Charles~L Isbell~Jr, and Andrea~L Thomaz.
\newblock Exploration from demonstration for interactive reinforcement
  learning.
\newblock In {\em Proceedings of the 2016 international conference on
  autonomous agents \& multiagent systems}, pages 447--456, 2016.

\bibitem{mujoco}
Emanuel Todorov, Tom Erez, and Yuval Tassa.
\newblock Mujoco: A physics engine for model-based control.
\newblock In {\em 2012 IEEE/RSJ International Conference on Intelligent Robots
  and Systems}, pages 5026--5033. IEEE, 2012.

\bibitem{fresh}
Baicen Xiao, Qifan Lu, Bhaskar Ramasubramanian, Andrew Clark, Linda Bushnell,
  and Radha Poovendran.
\newblock Fresh: Interactive reward shaping in high-dimensional state spaces
  using human feedback.
\newblock {\em Proceedings of the 19th International Conference on Autonomous
  Agents and MultiAgent Systems}, 2020.

\bibitem{trust1}
Zahra Zahedi, Sarath Sreedharan, Mudit Verma, and Subbarao Kambhampati.
\newblock Modeling the interplay between human trust and monitoring.
\newblock In {\em Proceedings of the 2022 ACM/IEEE International Conference on
  Human-Robot Interaction}, pages 1119--1123, 2022.

\bibitem{trust2}
Zahra Zahedi, Mudit Verma, Sarath Sreedharan, and Subbarao Kambhampati.
\newblock Trust-aware planning: Modeling trust evolution in longitudinal
  human-robot interaction.
\newblock {\em arXiv preprint arXiv:2105.01220}, 2021.

\end{thebibliography}

\end{document}